%% file: sample-sigconf.tex
\documentclass[sigconf]{acmart}

\usepackage{booktabs} 
\usepackage{multirow}
\usepackage{caption, subcaption}
\usepackage{adjustbox}
\usepackage{tabularx}
\usepackage{color, colortbl}
\definecolor{HighLight}{rgb}{0.88,1,1}
\definecolor{Gray}{gray}{0.9}
\renewcommand\footnotetextcopyrightpermission[1]{} 
\setcopyright{none}

\begin{abstract}
Multimedia applications often require concurrent solutions to multiple tasks. These tasks hold clues to each-others solutions, however as these relations can be complex this remains a rarely utilized property. When task relations are explicitly defined based on domain knowledge multi-task learning (MTL) offers such concurrent solutions, while exploiting relatedness between multiple tasks performed over the same dataset. In most cases however, this relatedness is not explicitly defined and the domain expert knowledge that defines it is not available. To address this issue, we introduce Selective Sharing, a method that learns the inter-task relatedness from secondary latent features while the model trains. Using this insight, we can automatically group tasks and allow them to share knowledge in a mutually beneficial way. We support our method with experiments on 5 datasets in classification, regression, and ranking tasks and compare to strong baselines and state-of-the-art approaches showing a consistent improvement in terms of accuracy and parameter counts. In addition, we perform an activation region analysis showing how Selective Sharing affects the learned representation.
\end{abstract}

\begin{document}
	
\title{Learning Task Relatedness in Multi-Task Learning\\ for Images in Context}

\author{Gjorgji Strezoski}
\affiliation{%
	\institution{University of Amsterdam}
	\city{Amsterdam}
	\state{Netherlands}
}
\email{g.strezoski@uva.nl}

\author{Nanne van Noord}
\affiliation{%
	\institution{University of Amsterdam}
	\city{Amsterdam}
	\state{Netherlands}
}
\email{n.j.e.vannoord@uva.nl}

\author{Marcel Worring}
\affiliation{%
	\institution{University of Amsterdam}
	\city{Amsterdam}
	\state{Netherlands}
}
\email{m.worring@uva.nl}

\maketitle

\input{samplebody-conf-acm_mm}

\begin{acks}
	The authors would like to thank Pascal Mettes for his feedback. This research is supported by the VISTORY project NWO award number 628.007.004.	
\end{acks}

\bibliographystyle{ACM-Reference-Format}
\bibliography{sample-bibliography}

\appendix

\section{Supplemental Material}

In this supplemental material we present an expansion of the experimental design and the results obtained from it. We reproduce some content of the main paper to make this supplemental more self-contained.

\subsection{Architectural Details}

\begin{figure}[!hb]
	\includegraphics[width=\linewidth]{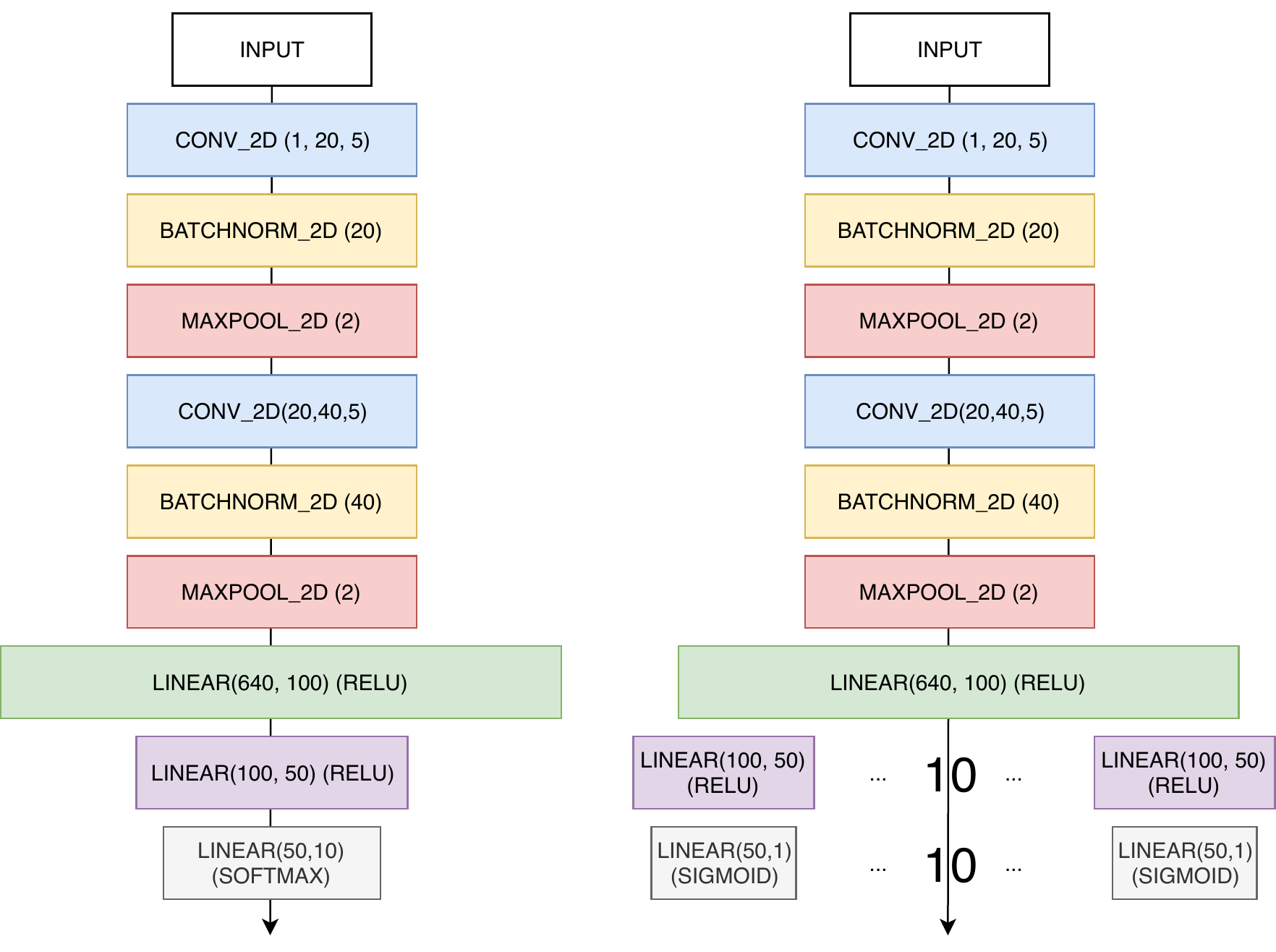}
	\caption{MNIST STL Model (left) and MNIST MTL Model (right)\label{fig_mnist_mtl}}
\end{figure}

\begin{figure}[!hb]
	\includegraphics[width=\linewidth]{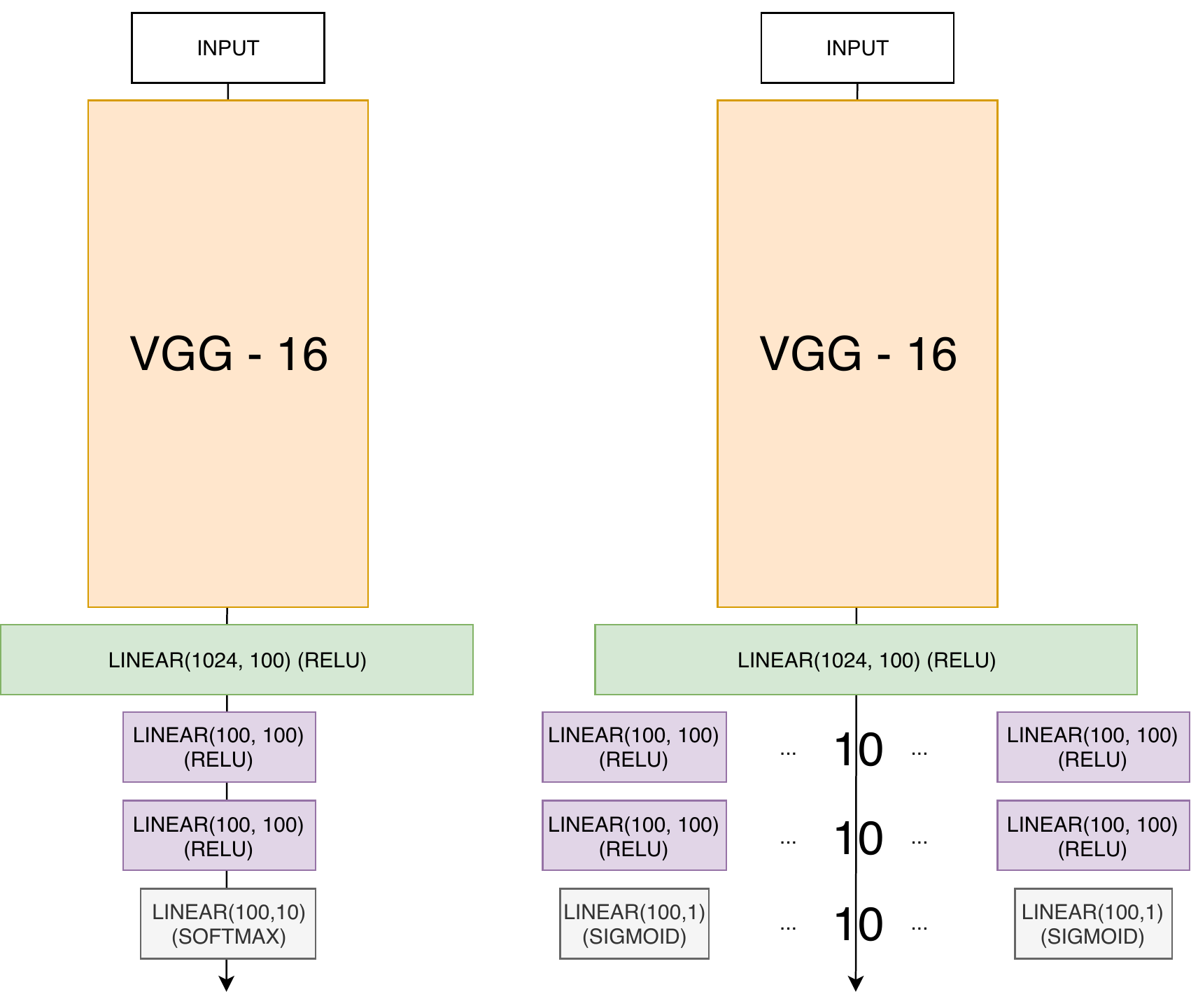}
	\caption{CIFAR STL Model (left) and CIFAR MTL Model (right)\label{fig_cifar_all}}
\end{figure}

\begin{figure}[!hb]
	\includegraphics[width=\linewidth]{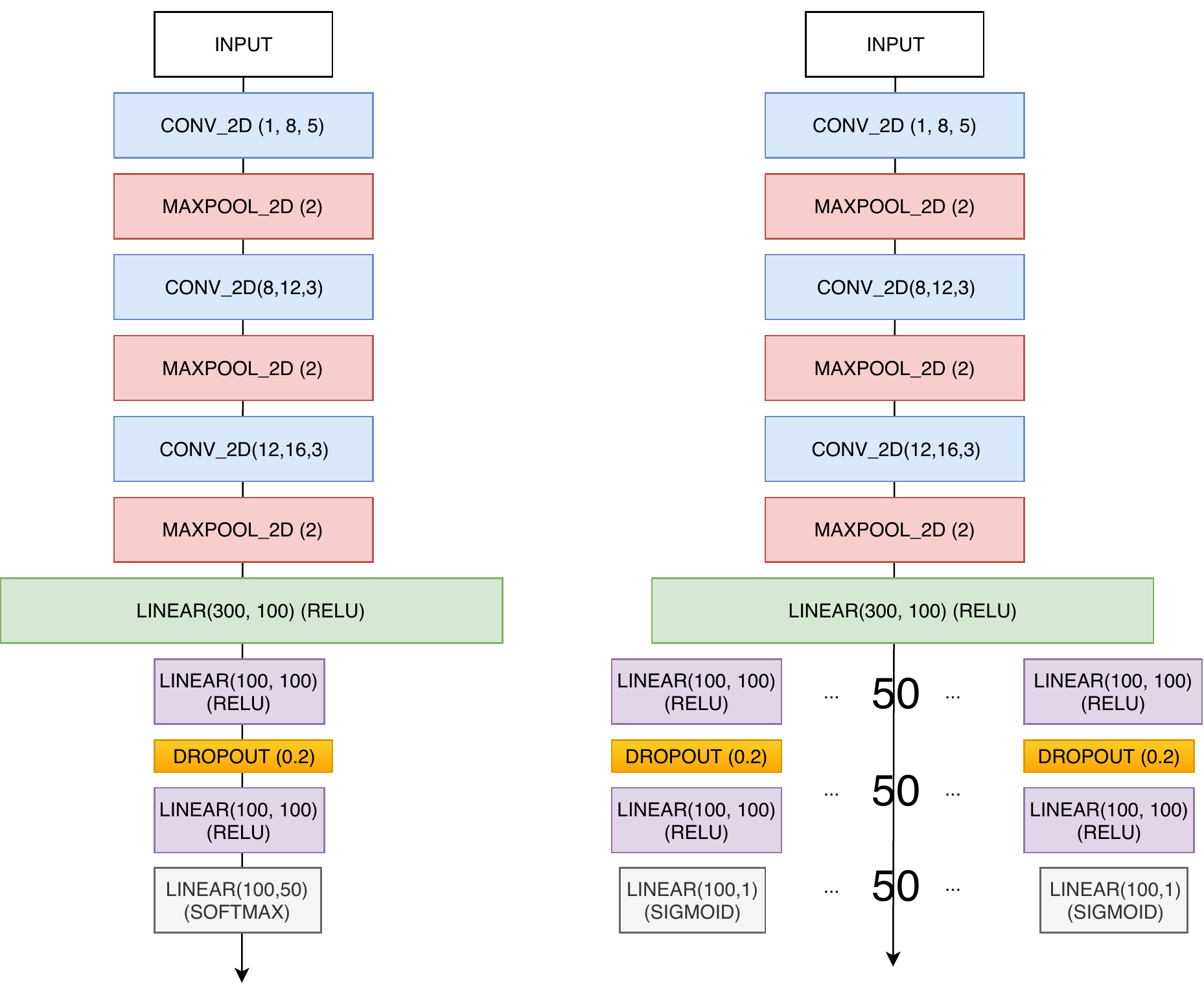}
	\caption{OmniGlot STL Model (left) and OmniGlot MTL Model (right) Diagram\label{fig_glot_stl}}
\end{figure}

\begin{figure}[!hb]
	\includegraphics[width=\linewidth]{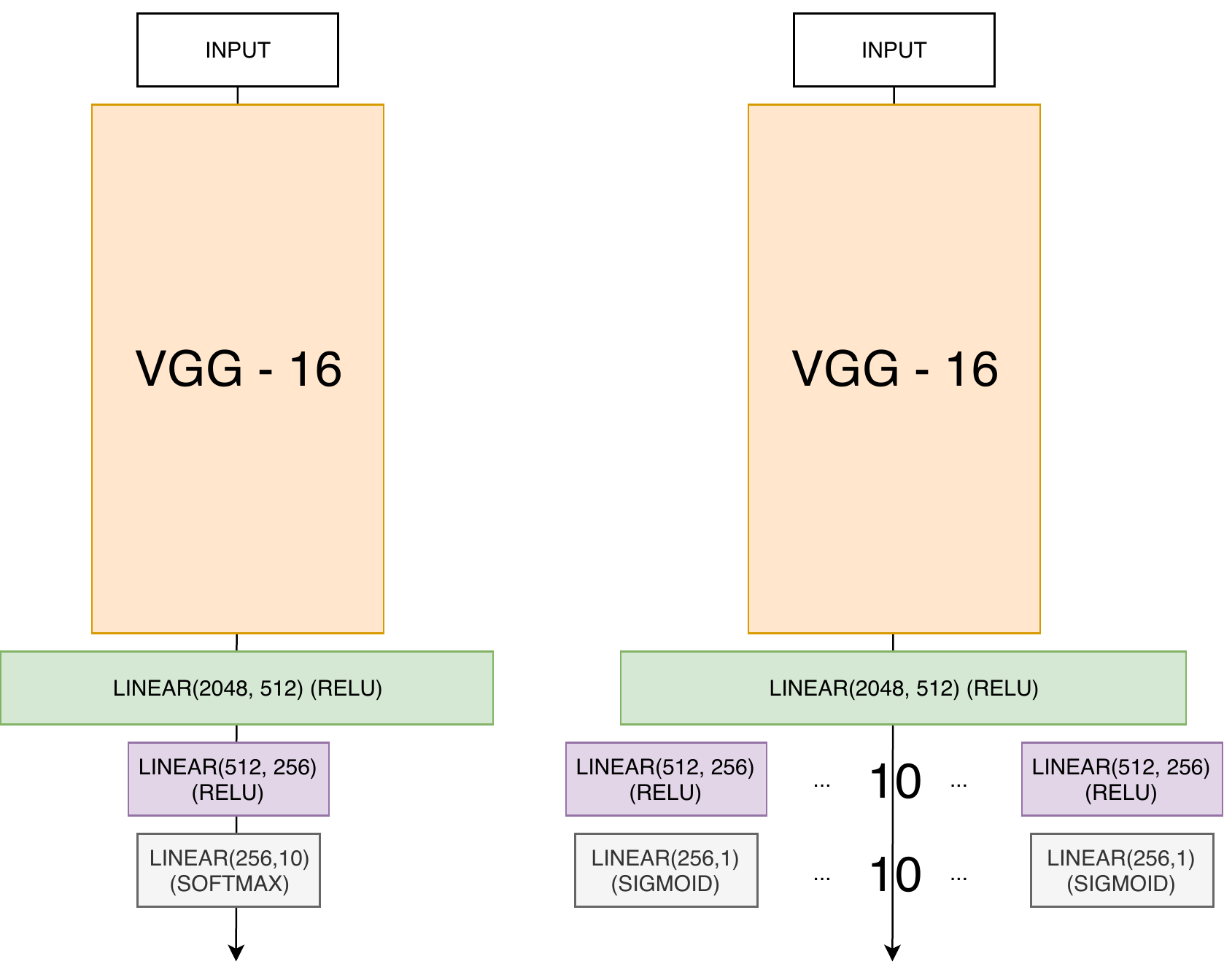}
	\caption{UCSD-Birds STL Model (left) and UCSD-Birds MTL Model (right)\label{fig_birds_mtl}}
\end{figure}

\begin{figure}[!hb]
	\includegraphics[width=\linewidth]{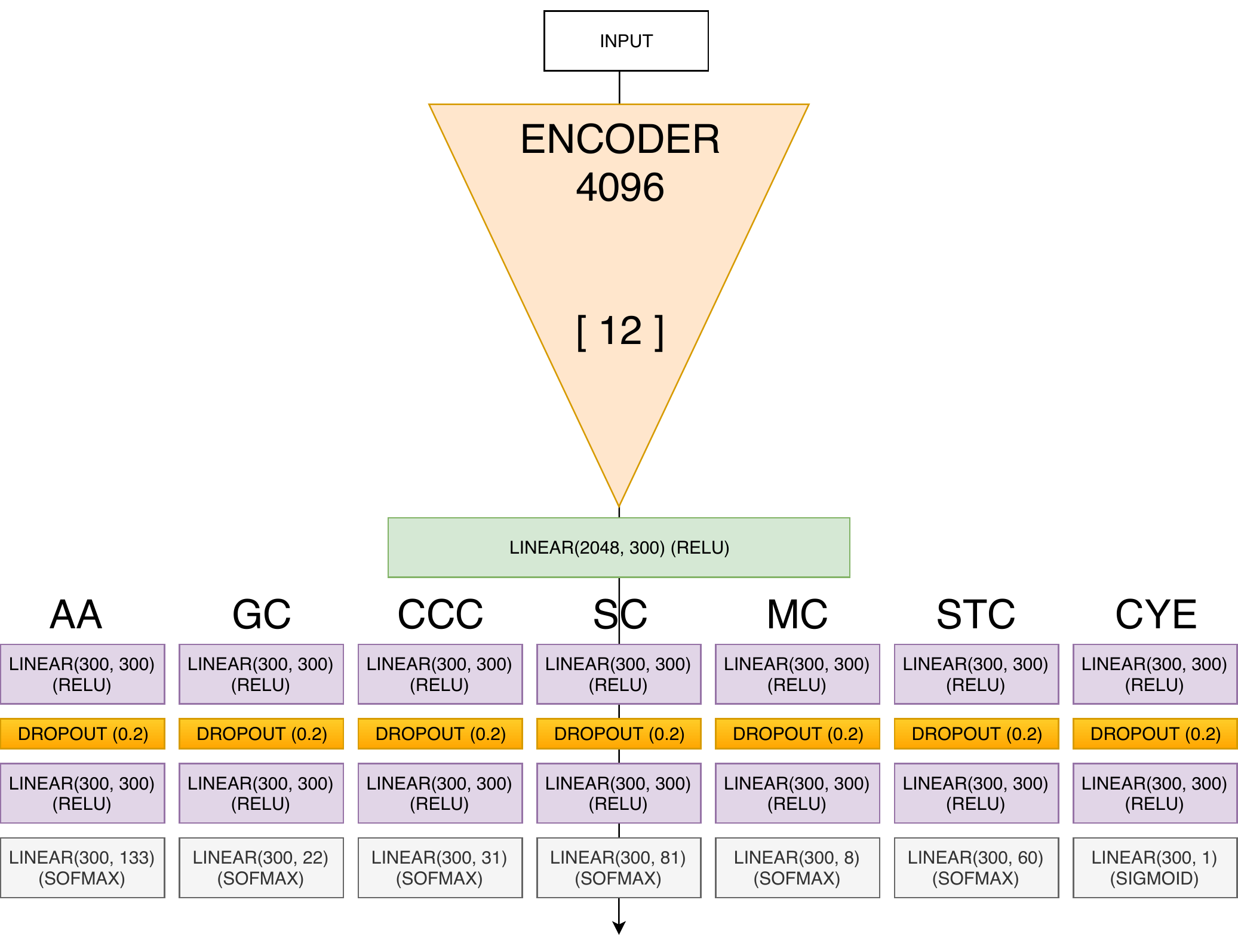}
	\caption{OmniArt MTL Model Diagram\label{fig_omniart_mtl}}
\end{figure}

This section contains a graphical overview of the defined models for easy understanding and reproduction of the experimental pipeline as well as tabular hyper-parameter definition. On the left side of Figures \ref{fig_mnist_mtl}, \ref{fig_cifar_all}, \ref{fig_glot_stl} , \ref{fig_birds_mtl} and \ref{fig_omniart_mtl} we depict the STL model, and on the right the MTL counterpart. The number between the task specific estimators is the amount of times the estimator is duplicated and each block is color coded by its functionality:

\begin{itemize}
	\item Yellow - BatchNormalization
	\item Orange - A complete feature extraction block or known architecture
	\item Dark Orange - DropOut
	\item Red -  2D Max Pooling
	\item Blue - 2D Convolutions
	\item Green - Shared Layers (Linear Fully Connected)
	\item Purple - Task Specific Linear Fully Connected Layers
	\item Grey - Task specific output layer
\end{itemize}

In Table \ref{tab_ssn_hyperparams} we show the complete hyperparameter and data setup for the defined STL and MTL experiments. All experiments on MNIST, CIFAR10, OmniGlot, OmniArt and Birds were run on a single nVidia Titan-X GPU with PyTorch version 0.3.1 on CUDA 7.5.

\begin{table*}[t]
	\centering
	\caption{A hyperparameter and setup table for the experiments. Input dimensions of 224px were achieved by center cropping over 256x images.}
	\label{tab_ssn_hyperparams}
	\begin{adjustbox}{max width=\linewidth}
		\begin{tabular}{lccccccccc}
			\hline
			Dataset    & Dataset Size           & Split 				   & \# Tasks           & Input Size (px) & \multicolumn{1}{c}{Task Type}   & Batch Size          & \multicolumn{1}{c}{Optimizer} & Learning Rate         & Momentum               \\
			\hline
			MNIST      & 70000 &\multirow{5}{*}{80-10-10} & \multirow{2}{*}{10}&28px & \multirow{3}{*}{Classification} & \multirow{4}{*}{64} & \multirow{4}{*}{SGD+Momentum} & \multirow{4}{*}{0.02} & \multirow{4}{*}{0.5} \\
			CIFAR-10   &      60000                  & 						   &                    &224px &                                 &                     &                               &                       &                        \\
			OmniGlot   & 1623                   &						   & 50                 &28px &                                 &                     &                               &                       &                        \\
			UCSD-Birds & 11788                  &					       & 10                 &224px & Ranking                         &                     &                               &                       &                        \\
			OmniArt    & 133000                 &						   & 7                  &224px & Classification, Regression      & 96                  & ADAM                          & 0.001                 & N/A                   
		\end{tabular}
	\end{adjustbox}
\end{table*}

\subsection{Extended Set of Results}

In this section we discuss the obtained results with multiple sharing conditions, the effects Selective Sharing has on training duration, the reason for the early stopping mechanism and the possibility of further refining the clustering process. Our expanded result set on multiple datasets provides insight into how the sharing conditional affect performance in different data types.

While in the main paper we report only the best obtained results (from Sharing with Similarity), we performed analysis on Selective Sharing with 4 sharing conditionals:

\begin{itemize}
	\item Sharing with similarity
	\item Sharing with dissimilarity
	\item Sharing with variance
	\item Sharing with random assignment
\end{itemize}

Each of these conditionals has a different effect on the learning process and produces different representations based on the groups of tasks being learned.

\begin{figure}
	\includegraphics[width=\linewidth]{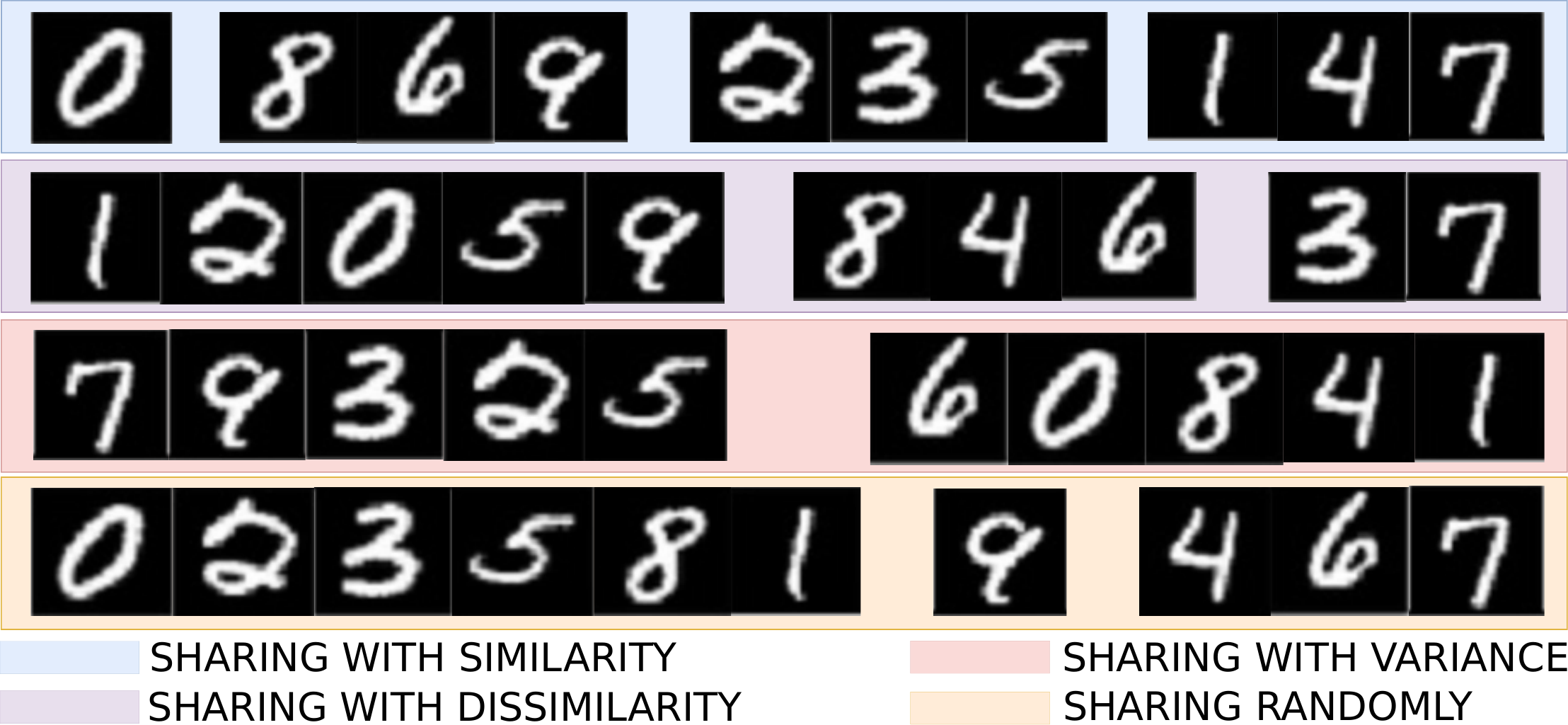}
	\caption{Task groups on the MNIST dataset with respect to the sharing criteria. In the similarity approach (blue) we cluster on the minimum core distance between the stable clusters, the dissimilarity (purple) approach maximizes the core distance so that we emphasize the sample diversity, the variance (red) approach defines clusters by the number of different task gradient factors near a point and the random (orange) approach is randomly picking a set of clusters. \label{fig_sharing_options}}
\end{figure}

In the case of sharing with similarity, a visual inspection of the input images shows that there is a correlation between the artist, the creation century and their school of origin. E.g. the works of Rembrandt and Modigliani in the OmniArt dataset have a oil and canvas in almost 80\% of the cases, which implies that knowing the artist narrows down the list of mediums significantly and vice versa. A similar conclusion can be deduced from sharing with similarity in the MNIST, OmniGlot and CIFAR-10 datasets. In MNIST the group containing '8', '6' and '9' shares the closed circle in each digit, while the group containing '2','3','5' is sharing the open loop on the bottom of '5', top of '2' and mid-section of '3'.

\begin{table*}[!t]
	\centering
	\caption{Performance comparison between a baseline STL approach, a baseline MTL approach and three Selective Sharing strategies on the OmniArt dataset. For every attempt we list the final number of trainable parameters, which for the STL approach is the sum of all the trainable parameters for all tasks. Since the feature extraction is fixed, the parameter count contains only the task specific branches and the hard shared layer. Best overall performance is achieved by Selective Sharing with similarity.}
	\label{tab_omniart_perf_extended}
	\begin{tabularx}{\textwidth}{lcccccccccc}
		\hline
		Task & AA    & GC    & CCC   & SC   & MC   & STC   & CYE  & \multirow{2}{*}{\# Params} & \multirow{2}{*}{Lock Epoch} & \multirow{2}{*}{\# Branches} \\ 
		Metric              		 	 	  & \multicolumn{6}{c}{\textit{Accuracy (\%)}}  & \textit{MAE (years)} 	       &  &     &  \\ \hline
		STL Baseline        	     		  & 28.0 & 25.0 & \textbf{32.1} & 24.7 & 41.3 & 19.2 & \multicolumn{1}{c}{144.32} & 3,487,664 & N/A & N/A\\
		MTL Baseline                 		  & 31.0 & 26.2 & 27.7 & 23.0 & 42.0 & \textbf{22.1} & \multicolumn{1}{c}{135.43} & 1,630,664  & N/A & 7  \\
		\rowcolor{Gray}
		\textbf{Selective Sharing - Similarity}       & \textbf{33.8} & \textbf{28.0} & 29.0 & \textbf{27.6} & \textbf{44.1} & 21.2 & \multicolumn{1}{c}{\textbf{128.11}} & \textbf{908,264} & 10 & \textbf{3} \\
		Selective Sharing - Dissimilarity     & 39.40 & 21.10 & 28.10 & 23.82 & 42.86 & 19.01 & \multicolumn{1}{c}{137.05} & \textbf{727,664} & \textbf{7}  & \textbf{2} \\		
		Selective Sharing - Variance 		  & \textbf{43.73} & 26.85 & 31.20 & 26.72 & 40.43 & \textbf{22.25} & \multicolumn{1}{c}{129.79} & \textbf{727,664} & \textbf{7}  & \textbf{2} \\
	\end{tabularx}
\end{table*}

\begin{figure*}[!h]
	\begin{minipage}[t]{0.45\textwidth}
		\includegraphics[width=\linewidth]{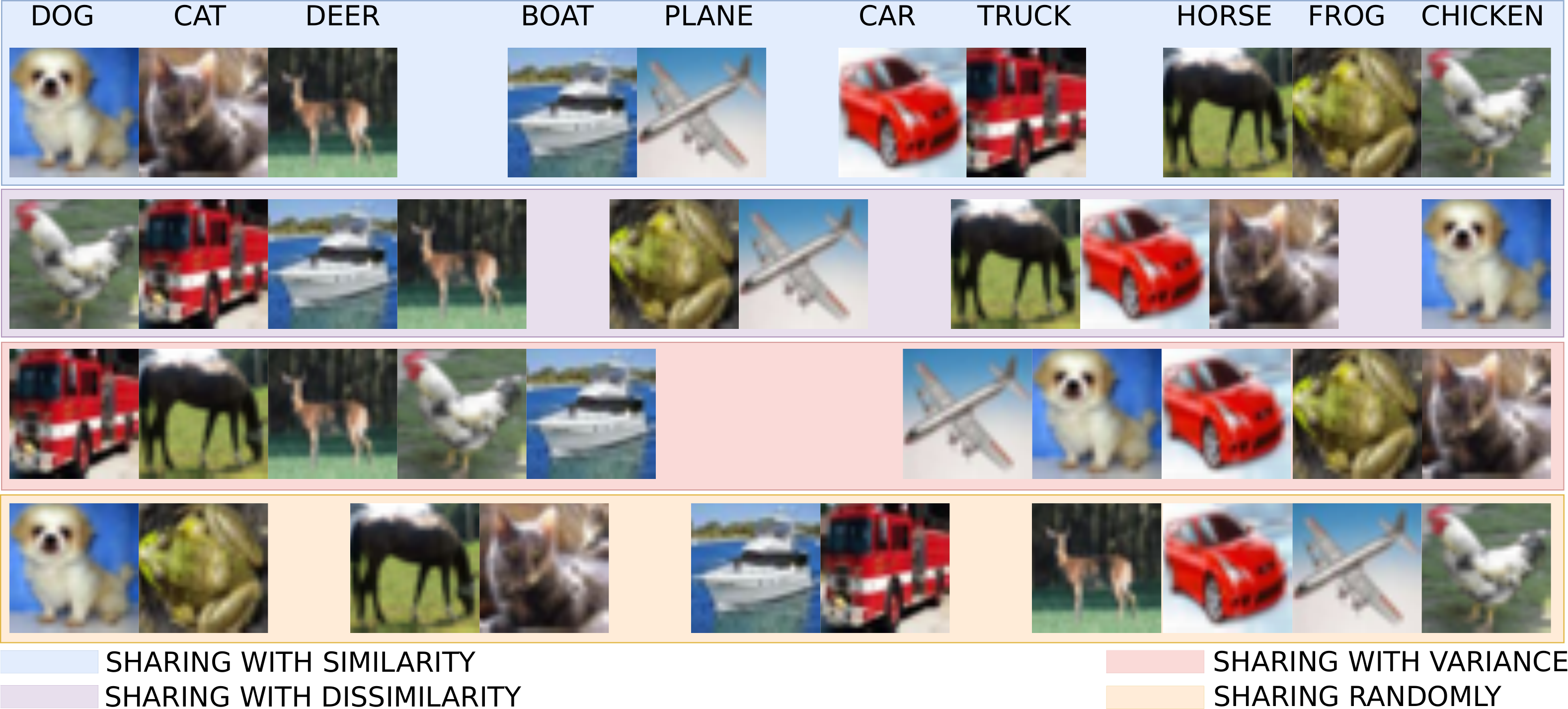}
		\caption{Task groups on the CIFAR10 dataset with respect to the sharing criteria. Upon inspection of the groups we can establish a link between animals regardless of the group formation conditional. Deer photographs are often in nature in a green scenery which might relate to the green color of frog skin or the surroundings of chickens. 
			\label{fig_sharing_options_cifar}}
	\end{minipage}
	\hspace{1.3cm}
	\begin{minipage}[t]{0.45\textwidth}
		\includegraphics[width=\linewidth]{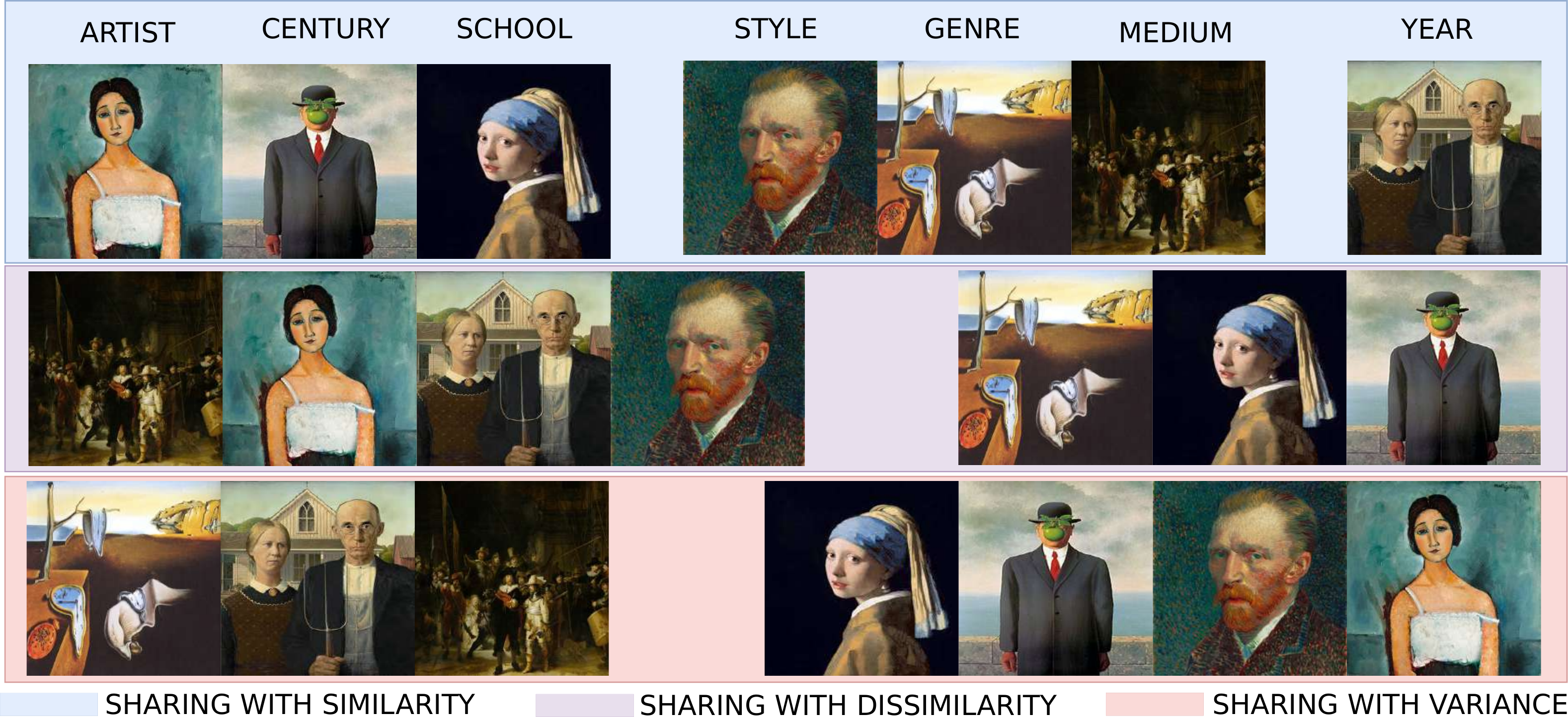}
		\caption{Task grouping with Selective Sharing on OmniArt. Best overall performance is achieved with sharing with similarity. The STC, GC, MC group experiences a performance boost when learned together. \label{fig_sharing_options_omniart}}
	\end{minipage}
\end{figure*}

\begin{figure*}[!h]
	\includegraphics[width=\textwidth]{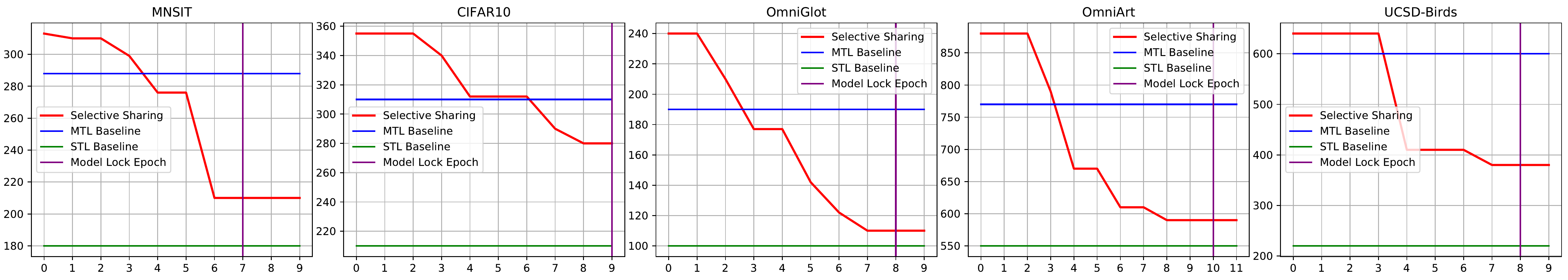}
	\caption{Epoch duration comparison between STL, MTL and best performing Selective Sharing approach.\label{fig_durations}}
\end{figure*}

\begin{table}[]
	\centering
	\caption{Extended performance comparison for MNIST, CIFAR10 and OmniGlot for Selective Sharing, an STL and MTL baseline. Prior knowledge for relatedness between the 50 OmniGlot alphabets was missing so a fair comparison is not possible.}									
	\label{tab_mnist_cifar_omniglot_ext}
	\begin{adjustbox}{max width=\linewidth}
		\begin{tabular}{lccc}
			\hline
			Dataset              					 	& MNIST  			& CIFAR10 		  & OmniGlot     \\
			Metric               					 	& Accuracy (\%) 	& Accuracy (\%)   & Error Rate   \\
			\hline
			STL Baseline 		& 98.0 			&  92.6		      & 0.34         \\
			MTL Baseline         						& 96.8 			&  90.3 		  & 0.29        \\
			MTL Stitching 	 	& 98.4 			&  92.1 		  & n/a          \\
			MTL MSD 	  	 		 	& 98.7 			&  91.9 		  & n/a          \\
			\rowcolor{Gray}
			\textbf{Selective Sharing - Similarity}  	& \textbf{99.0} 	&  \textbf{92.7} &\textbf{0.24} \\ 
			Selective Sharing - Dissimilarity  	& {96.9} 	&  90.1 & 0.33 \\ 
			Selective Sharing - Variance 	& 98.6	&  91.0 &0.28 \\ 
			Selective Sharing - Random 	& 94.2	&  91.1 & 0.28 \\ \hline  
		\end{tabular}
	\end{adjustbox}
\end{table}

For sharing with between cluster variance, the results show that for OmniArt it is better to build a representation with respect to a larger number of tasks rather than one formed from smaller task groups. This phenomenon can be attributed to the tasks initially having a high correlation. E.g. individual artists have contributed to a limited amount of styles and periods, on chosen mediums, from their own school of painting, which encapsulates their entire descriptive metadata. Also it is quite intuitive that sharing with variance ensures the learned representation will encompass as much relevant information as possible.

Sharing with dissimilarity is motivated by the fact that an increased information entropy in the feedback signal would produce more robust and general features. Opposite of sharing with similarity, we train a smaller more general representation for the group of dissimilar tasks. If we observe the group formed by 'Frogs' and 'Airplanes' in the CIFAR-10 experiment, this branch of the network would have learned a more general representation then the 'Boat' and 'Airplane' group. Just by being exposed to images with different contexts, in theory, improves the overall generalization properties of the branch.

We use sharing randomly as a control signal to which we compare performances, it uses randomly generated sequences to select clusters and then checks if a task is duplicated (duplicates are removed). A constant improvement over this method in all cases further confirms the beneficial effects of selective sharing in MTL models.

This threshold can be a either a user defined, or a trainable parameter. A user defined threshold would imply that prior knowledge has been introduced to the system which no longer makes Selective Sharing a \textit{no prior knowledge} MTL method. However, there a ways to monitor and learn the performance threshold by following individual and group task performances during training.

In our case, our empirical study shows that it is best to start applying Selective Sharing in the first epochs after the model starts learning. At this point in time the gradients are more stable and consistent, which acts as a regularizer for the extracted factors and improves our clustering performance. If Selective Sharing is applied very late in the training process, the clustering mechanism should be much more sensitive to compensate for the minute differences in the gradient factors.

\subsection{Training duration with Selective Sharing}

Selective sharing involves additional steps during the training procedure, this is the factorization and clustering processes. While in the start of the training process it increases the duration of an epoch, as the model reduces in parameter space epoch duration decreases multiple times. After the early stopping mechanism takes over and a final model architecture is defined, epoch duration stabilizes and is always lower than conventional MTL approaches.

Compared to a signle STL model performing one task over a 10-way softmax, MTL will always be slower, and so will Selective Sharing. 

\end{document}

%% file: samplebody-conf-acm_mm.tex
\section{Introduction}

In many multimedia applications, there are not only multiple sources of information, but a multitude of tasks to perform as well. Whether segmenting sketches of animals \cite{sarvadevabhatla2017sketchparse}, recognizing birds by attribute appearance \cite{mtl_msd}, attributing artists to paintings \cite{DBLP:journals/corr/abs-1708-00684} or classifying facial features \cite{gnas_acmmm18, Liu_2018_CVPR}, state-of-the-art results \cite{mtl_msd, gnas_acmmm18, Liu_2018_CVPR} show that exploiting task relatedness in a multi-task learning (MTL) setting benefits performance. For example, an artwork can be described by its creation period, style, technique, genre, type, or artist, as well as with content descriptors such as brush-stroke frequency, lighting direction or repeated texture patterns. Learning to recognize or predict each of these contextual attributes is a separate task, the solution of which may benefit the other tasks in the pool. This way, if one of the attributes is missing, or unknown, the knowledge from learning other contextual attributes, both visual and textual, can help in narrowing down the space of potential values for the missing attribute. In other words, if we recognize a painting has been made in the 17th century in the Dutch school of painting and contains a gray cloudy landscape, in an artist attribution task we could short-list a few artists as the possible creator. 

The main benefit of MTL is leveraging multiple shared information sources, especially when a mutual dependence is present to improve the process of solving multiple tasks at once \cite{li2017multi, sarvadevabhatla2017sketchparse, he2017adaptively}. While theoretically the gain from MTL is enticing, fully exploiting its benefits is a tedious process in practice. Often this is due to the lack of expert knowledge about the data posing the MTL challenge. Knowing how to setup the shared representation in a MTL setup can play a key difference in the model's performance. However, a recurring theme in existing MTL approaches is that the task relatedness still needs to be specified before training based on prior knowledge. Noticing task relatedness comes naturally to humans, but for MTL models to do the same a high level of domain expert knowledge is required which is often expensive and rarely available. In this work we propose a method - \textbf{Selective Sharing} - for learning task relatedness during training, and modifying the network architecture on-the-fly with no prior knowledge.

\begin{figure}
	\includegraphics[width=\linewidth]{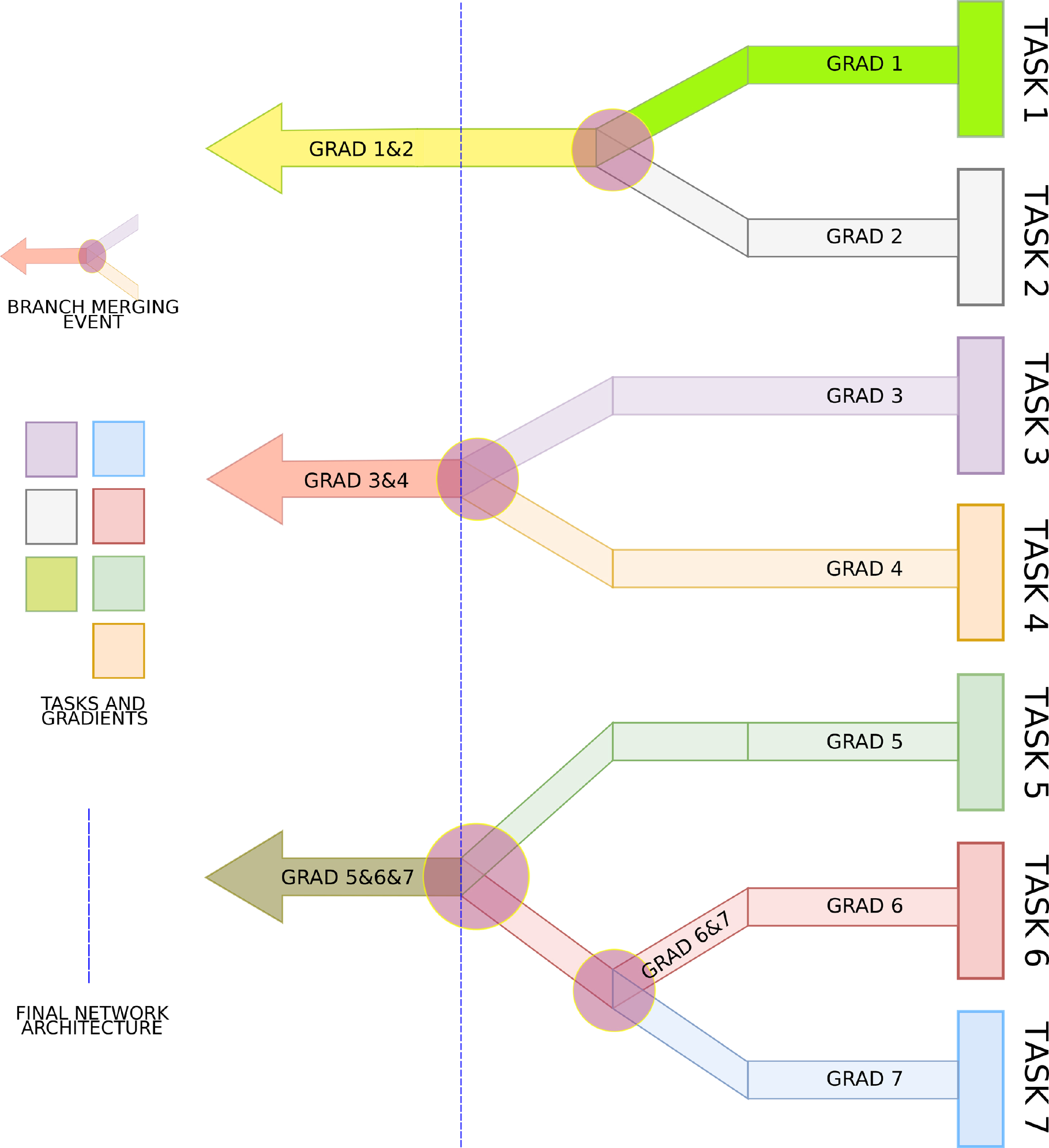}
	\caption{Illustrating the gradient flow of the individual task branches as they group through the training epochs with the simplifying assumption that all task specific branches are identical and identically initialized. \label{fig_high_level_grads}}
\end{figure}

Selective Sharing makes it possible to optimize multiple tasks at once by only sharing the weights and parameters between tasks. Sharing occurs when their relatedness satisfies an influence metric derived from latent secondary features, such as gradient factors. As visible in Figure \ref{fig_high_level_grads}, we measure task similarity during training and cluster tasks with similar responses to the same stimuli. Upon satisfying the metric these tasks form groups and continue training together.

This approach iteratively reduces the size of the parameter space compared to other MTL approaches without the requirement of a prior inter-task relationship specification. Grouping tasks allows them to share and fit a mutual parameter space. This reduces trainable parameter count in half per pair of tasks, once one of the task specific branches is no longer used. With a greedy starting position, the parameter reduction properties of Selective Sharing shorten training epoch duration and decrease memory requirements for the final model.  

In this paper we elaborate on the following three contributions:

\begin{itemize}
    \item  We propose Selective Sharing, a new approach for MTL that performs data driven task grouping without requiring prior knowledge.
    \item  We demonstrate our approach's ability to make explicit the latent task relatedness in MTL supported with task specific activation region analyses.
    \item  We improve or maintain performance in classification, regression, and ranking tasks while reducing the size of the parameter space.  
\end{itemize}
 
\section{Related Work}

MTL has developed significantly from the moment it was introduced \cite{Caruana1997, stein1956inadmissibility}. While originally it was a 'one size fits all' methodology, multiple MTL design patterns have been identified depending on the type of data that is being modeled, the type of sharing between the different tasks or the different levels on which a mutual representation is created. Additionally, with MTL comes a natural urge to simplify the models at hand and group the tasks that would benefit each other's learning process. With this mutually beneficial task relationship in mind, there are numerous domains and modalities \cite{taskonomy2018, 2017arXiv170508142R, li2017multi, sarvadevabhatla2017sketchparse, he2017adaptively, jou2016deep, wang2016human, kaneko2016adaptive} where the MTL methodology can be applied. As such, MTL is often used implicitly without a specific reference in methods such as transfer learning and fine-tuning \cite{Caruana1997, sharif2014cnn} as well. Given this broad scope of MTL in this section we focus on MTL in the context of systems that exploit visual information with their accompanying metadata.

\subsection{Multi-task Learning} 

Whenever we find ourselves optimizing for more than one loss function, we are essentially doing MTL. This is in contrast to single task learning (STL) where we optimize only one loss function \cite{DBLP:journals/corr/Ruder17a}. Caruana in 1997 also defined it as training tasks in parallel while using a shared representation \cite{Caruana1997}. While this is a simple definition, it does cover a vast portion of the possible problem space and allows for a lot of freedom in its interpretation.

In MTL there are two general ways in which we can define a problem and several more ways to apply an MTL approach with respect to the nature of the problem. If we are going to label an MTL problem by whether the tasks are sharing a label space or not, it can be either a Homogeneous MTL problem or a Heterogeneous MTL problem.

\subsubsection{\textbf{Homogeneous and Heterogeneous MTL}}

Homogeneous MTL problems are usually defined over existing multiclass classification problems that do not naturally consist of multiple tasks. This definition allows us to study the effects of MTL in a controlled environment, often competing against strong specialized STL benchmarks. Massimiliano and Theodoros in \cite{evgeniou2004regularized} demonstrate a homogeneous formulation by creating 139 binary tasks predicting whether a pupil belongs to a school or not. In a visual domain the same can be achieved with the  the USPS dataset \cite{hull1994database} as Kumar et al. show in \cite{kumar2012learning}. Yang and Hospedales \cite{yang2016deep} make use of the homogeneous formulation over the MNIST dataset \cite{lecun1998gradient}. In summary, with a homogeneous definition the target space is transformed to fit the MTL paradigm by splitting each target into a separate \textit{one vs. all} task. 

Heterogeneous MTL primarily occurs in problems and datasets that contain multiple types of labels existing in separate spaces for their data \cite{yang2009het}. For example, an artwork can have a label for the artist, the creation period, the type and materials of the artwork. All of those labels represent different tasks like artist attribution or creation period estimation \cite{DBLP:journals/corr/abs-1708-00684}. Zhang et al \cite{zhang2014facial} demonstrate the same for facial landmarks and Zamir et al. \cite{taskonomy2018} show a heterogeneous formulation over multiple domains and target spaces at the same time.

Regardless of the problem formulation, following Caruana's definition of MTL we know that a shared representation is key. But how do we decide what to share, to which extent and between which tasks should the sharing occur?

\subsubsection{\textbf{Hard and Soft Parameter Sharing}}

Most MTL approaches share the same base structure for feature extraction \cite{gnas_acmmm18, mtl_msd, zhang2014facial, liu2015multi, zhang2016deep, evgeniou2004regularized, mrkvsic2015multi, DBLP:journals/corr/Kokkinos16, taskonomy2018, liu2015representation} and then continue to branch out, intertwine or widen the model's parameter space. Sharing is an essential part of MTL and can be categorized as hard sharing or soft sharing.

Hard parameter sharing is the most commonly used approach for MTL in neural networks and goes back to \cite{caruana1998multitask}. It is generally applied by sharing the hidden layers between all tasks, while keeping several task-specific output layers. Hard parameter sharing greatly reduces the risk of over-fitting. In fact, Baxter \cite{Baxter1997} showed that the risk of over-fitting the shared parameters is order N (where N is the number of tasks) smaller than over-fitting of the task-specific parameters, i.e. the output layers. This makes sense intuitively: the more tasks we learn simultaneously, the more our model has to find a general representation that is suited for all of the tasks and our chance of over-fitting on our original task is smaller.

In contrast, soft parameter sharing each task has its own model with separate parameters. This means that there is no physical layer structure that is shared between the different tasks, but rather it is an intermediate set of parameters that is shared. In this case, sharing parameters acts as a weight regularizer as Duong et al. show in \cite{duong2015low}. Further, Yang et al. in \cite{yang2016trace} show that soft sharing regularization works on factorized representations (Tensor-Train \cite{oseledets2011tensor} or Tucker \cite{tucker1966some}) as well. This kind of sharing promotes task independence, but keeps the parameter space similar so that tasks can still influence each other. Similarly, Misra et al. \cite{misra2016cross} introduced Cross-Stitch Networks that rest on the soft sharing paradigm. Initially their approach starts with identical structures between tasks and soft parameter sharing. Different from \cite{yang2016trace} and \cite{duong2015low}, in this model the sharing is determined by cross-stitch units, placed after pooling or fully connected layers, whose task is to learn a linear combination of the output of the previous layers from both structures. A shortcoming of these approaches is that they require knowledge of the parametric form of the tasks at hand. In \cite{mtl_msd}, Mejjati et al. treat their tasks as random variables for which statistical dependence can be measured and maximized. While it achieves comparable performance on several ranking and regression tasks, their approach relies on an existing dataset that mimics the distribution of the tasks. In this way, for this method to obtain the reported performance, prior knowledge of the distributions for each task is always necessary. Inspired by \cite{mtl_msd}, in Selective Sharing we aim to obtain this knowledge on the fly, while training the model in a greedy fashion accumulating such information as it becomes available in the back-propagated gradients.

\subsubsection{\textbf{Task Grouping}}

Choosing which tasks share parts of their parameter space proves to be beneficial when prior knowledge about this entanglement is present. Long and Wang \cite{long2015learning} introduced the concept of Deep Relationship Networks where they impose matrix priors on the shared fully connected layers which allows the model to learn the task relationships. This is a similar method to Bayesian MTL models \cite{daume2009bayesian, marquand2014bayesian}, who essentially group the tasks based on this predefined relationship matrix. This approach to grouping still relies on a predefined sharing structure which we avoid using Selective Sharing. Liu et al \cite{liu2017distributed} propose a dynamic greedy bottom-up approach to task grouping to overcome the predefined sharing matrix problem. Instead of hard-coding task dependencies, they dynamically widen the model by creating new branches as training progresses. However, this method has a risk of every task becoming a separate branch in the structure and therefore limits sharing and parameter tuning between tasks. Other approaches have tried to generalize existing approaches to MTL like Yang et al. \cite{yang2016deep}, who uses tensor factorization to split the parameter space into shared and task specific parameters of every layer. Sluice networks introduced by Ruder et al. \cite{ruder2017sluice} aggregate multiple MTL approaches into one by creating a task hierarchy in MTL problems to maximize relatedness utilization regardless of how related they actually are. Regardless of the scenario, task grouping methods tend to rely on prior task relationship knowledge, or extensive statistical analysis prior to model design.

In this paper we address some of the issues that arise with deep MTL models. Namely, we stray away from the predefined sharing structure and let the model decide the sharing parties and their structure. This allows for flexibility and dynamically adapting models that tune themselves to the tasks at hand. Learning the underlying task relatedness based on the supervision responses over a shared input, allows for grouping tasks on the fly during training. The resulting network then provides additional insight (explicit task relations) into the data we are analyzing.

\section{Selective Sharing}

In any deep learning system, gradients flow from the final layers of the model towards the starting ones carrying a corrective signal for the weights and biases along the way. They are a way for the model to know where and how much it should correct its trainable parameters. Selective Sharing is based on the assumption that the identically constructed task specific estimators, sharing the same input and feature extraction platform, would manifest a correlation between the back-propagated gradients for related tasks. Following this, we define an MTL problem as:

\begin{itemize}
	\item A set of tasks $T$ with cardinality $|T|$.
	\item A set of related tasks $R_k=T/t_k$ for each task $t_k\epsilon T$.
	\item A set of task dependent loss functions  $E_t$ generating task specific gradients $G_t$ for the task specific targets $K_t$. 
	\item A set of identical task specific estimators with identical layers $l_t$ and weights $w_{l_t}$.\footnote{As all equations are task-specific, $t$ is omitted in further notation for simplicity and readability. Inputs, outputs, targets and activations follow the classic machine learning naming convention ($x, \hat{y}, y, a $).}
\end{itemize}

\begin{figure*}[!h]
	\includegraphics[width=\linewidth]{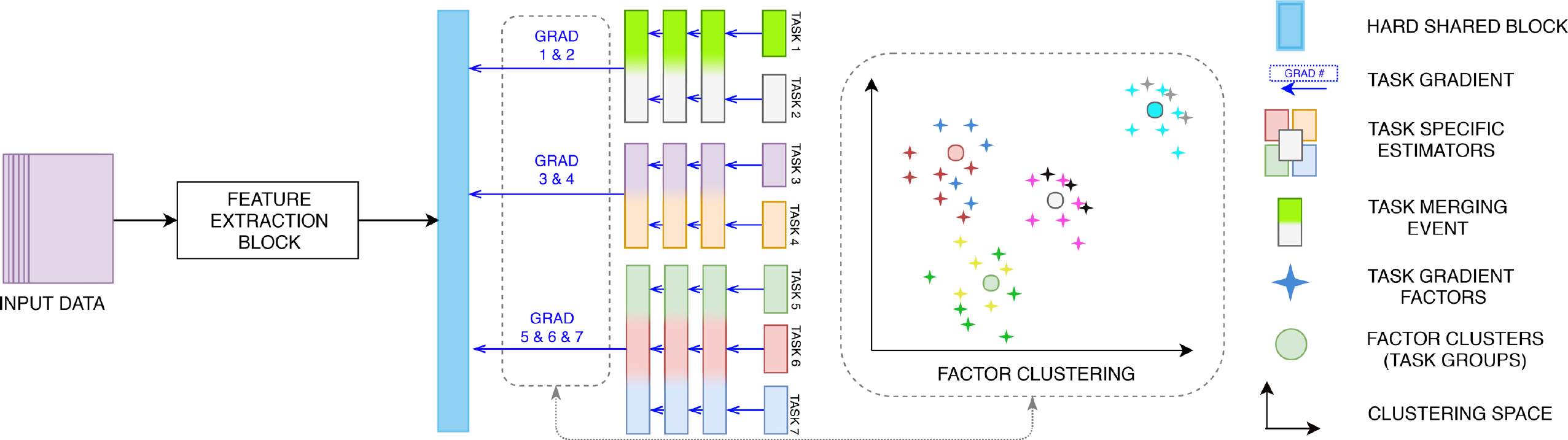}
    \caption{A high level overview of the Selective Sharing architecture. All tasks share the same input data (purple) and pass through the same feature extraction block before arriving to the shared representation (blue block). From the final shared representation all tasks branch in identical estimators with a different output (per task). With each mini-batch we accumulate gradients and factorize them. After clustering the factorized gradients from the set of tasks, the architecture is recomputed to group tasks whose gradients belong to the same factor cluster which implies keeping one estimator and discarding the rest so that the group tasks would only differ in their output. 
    \label{fig_high_level_overview_ss}}
\end{figure*}

Not having the constraint of requiring domain specific task relatedness knowledge, this approach allows for formulating any set of classification, regression or ranking problem into the MTL paradigm and can be used for discovering relations between contextual attributes in large bodies of data. For example, if we observe an MTL model trained for classifying written digits with one branch per digit, we can postulate that our optimization scheme will generate similar corrections for the branch specific weights in the branches related to classes '1', '4' and '7' (see Figure \ref{fig_sharing_options_all}), in the case where the input image is the number seven. Having a shared input and distinct gradient flows, we can study the gradients and their factors which depict the behavior of task specific estimators, divulging information about intertask relatedness and supervision signal similarity. In this way, we define three functional structures of our approach: the feature extraction block, the shared representation block and the task specific estimators (branches) as illustrated in Figure \ref{fig_high_level_overview_ss}.

The feature extraction block provides the initial representation for the input data. Depending on the problem and whether it should be trained or not, the feature extraction block can range in form from a CNN to SIFT, HOG or any other handcrafted features that apply. The features are then propagated forward through the hard shared layer, which is the base representation Selective Sharing is working with. From this point, task specific estimators branch out with identical blocks (identical layers $l_t$ with identical weights $w_{l_t}$) per task $t$ with $K_t$ targets, with a task specific loss function $E=\frac{1}{2}\sum_{k\in K} (\hat{y}_k-y_k)^2$ for example Mean Squared Error, $\hat{y}_k$ being the activation of output unit $k$ with input $x_k$, and $y_k$ as the ground-truth for the same units producing the task specific gradients $G_t$. 

\begin{figure}
	\includegraphics[width=\linewidth]{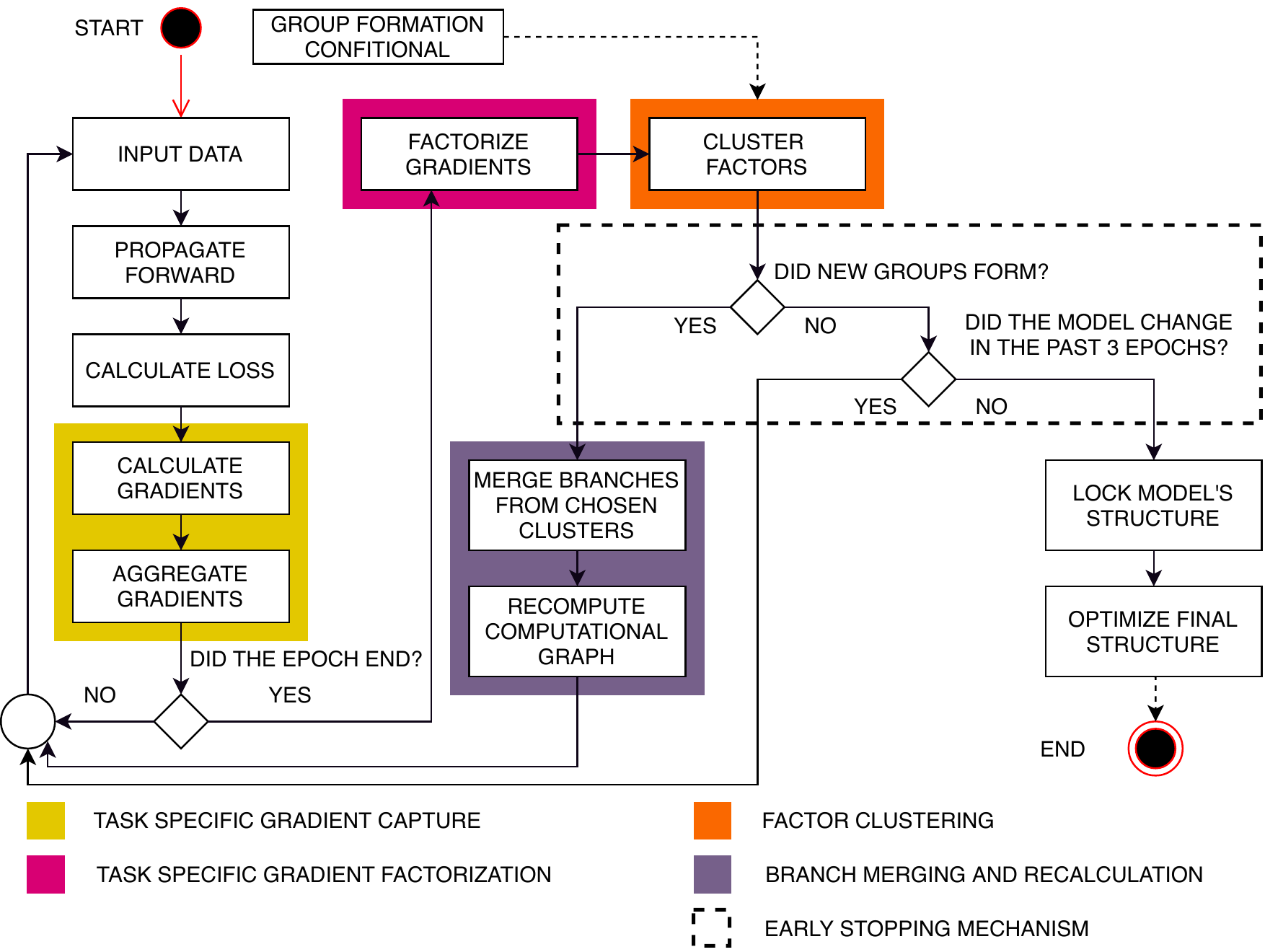}
    \caption{An operational flow diagram of the Selective Sharing method. The different stages in the method are color coded and the region enclosed by the dashed line represents the early stopping conditions. 
    \label{fig-selective-sharing-uml}}
\end{figure}

Our approach is applicable in both heterogeneous and homogeneous problems. For homogeneous MTL problems (MNIST, CIFAR, OmniGlot, Birds) we define each category as a binary classification task. Our model changes accordingly with duplication of the estimator for each task. In a heterogeneous (OmniArt) problem we optimize for multiple different estimators suitable for the task at hand. 

The name Selective Sharing is derived from an important property of the method itself, which is allowing the model to select the tasks where sharing should occur, without it being specifically programmed to do so. There is however, a general directive for the sharing. The directive is given before the model training starts and is essentially the clustering condition. Depending on the goal the model can perform sharing by:

\begin{itemize}
	\item \textbf{Similarity} maximizes the model's exposure to data samples which have a high probability of sharing similar primitive features or high level semantics.
	\item \textbf{Dissimilarity} allows for an increase in information entropy which can prolong training times, but results in better generalization.
	\item \textbf{Variance} favors large task clusters so it strives towards learning a more complete representation.
\end{itemize}  

\subsection{Task specific gradient capture}

The Selective Sharing pipeline starts by tapping into the gradient flow at the point where the model starts branching out for different tasks, immediately after the hard shared representation layer and task specific layer $l_{t}$. In that point we define the gradient as $G_t$ and compute it as a gradient for weights $w_{ij} \in l_{t}$ as follows (bias excluded for simplicity): 
	\begin{equation}
	G_t=\frac{\partial E_t}{\partial w_{ij}}=\sum_{k \in K_t}(\hat{y}_k - y_k)g'_k(x_k)\frac{\partial}{\partial{w_{ij}}}x_k,
		\label{eq_gradient_hidden_layer}
	\end{equation}
if we continue to decompose activation $x_k$ and define the gradient from the previous layer for task $t$ as $\delta_{tj}=g'_j(x_k)\sum_{k \in K_t}\delta_kw_{jk}$, the gradient gets the final form of:

	\begin{equation}
	G_t=\delta_{tj}a_i,
	\label{eq_gradient_hidden_layer_final}
	\end{equation}

Each $G_{t}$ tensor is a gradient from a separate task estimator. At the hard shared representation layer where $g_j$ is the activation function for node $j$ in layer $l_{t}$, $w_{ij}$ are the weights connecting the node $j$ in layer $l_{t}$ with the node $i$ in layer $l_{t}-1$ and $a_i$ is the activation or output for node $i$ in layer $l_t$. Using the same notation, we define $\delta_k$ as all the terms that involve the value of a unit at index $k$ with respect to the expected target $K_t$ and $z_k$ is the input to node $k$ of layer $l_t$ for task $t$.

Since these captured gradients represent the data being clustered on basis of task relatedness, we can consider them as secondary features to the ones we learn for the data representations. Because the gradient capture is performed in a fully connected section between the hard shared representation and task specific branches, the gradient affecting the hard shared layer is:

	\begin{equation}
		G = \sum_{t=1}^{|T|} G_t,
		\label{eq_gradient_sum}
	\end{equation}
	
When observing a gradient in a fully connected section of the model, there is an influence between each pair of input and output nodes. Despite their high dimensionality, we collect and label these gradient tensors making the resulting group tensor a clustering candidate and name it after the index of the task specific estimator (e.g. in our digit recognition model a gradient coming from estimator seven would have the label seven).

Keeping in mind that this capture is performed between each task branch and the hard shared representation layer a rather compact representation with little information loss is necessary in order to perform the task correlation analysis. Additionally, this process is spanning through hundreds of examples passing in an epoch and the aggregated information loss can be substantial. For this reason we utilize the Tensor Train decomposition \cite{oseledets2011tensor} which not only allows for a compact representation of tensors, but eases the application of linear algebra operations.   

\subsection{Task specific gradient factorization}

Depending on the dimensions (number of units) of the layers where they are calculated, gradients can become high dimensional tensors. This is particularly visible in MTL pipelines with multiple branches, as they all generate their own gradients. This type of rich representation is informative, but poses a problem in a setting where aggregation and quick analysis is required. Tensor train decomposition \cite{oseledets2011tensor} is a mechanism that transforms a given tensor $G_t$ with $d$ elements $G_t(f_1,f_2,..., f_d)$ into a more compact representation $B_t$:

	\begin{equation}
		B_T=\sum_{i=1}^{d}D_i(\alpha_{i-1}, f_i,\alpha_i),
		\label{eq_decomposition}
	\end{equation}
	
where each of the $D_i$ elements are matrices representing a \textit{factor} of the tensor $G_t$ and the $\alpha_f$ represent the index matrix of dimension (factor) $f$. If we aim to reconstruct the whole tensor a multiplication needs to be performed between all of the factors and then a summation of the auxiliary $\alpha$ index nodes. In a reconstruction setting these operations are performed for as many times as there are dimensions in the decomposed tensor. From them, branch specific matrices $C_t$ are produced with dimensions $[num\_batches \cdot shape(B_t)]$ where $t$ is the index of the task.

	\begin{equation}
		C_t=\begin{bmatrix}
			B_0  
			B_1 
			B_2 
			...
			B_{m}
		\end{bmatrix}^T,
		\label{eq_feature_matrix}
	\end{equation}
	
The $C_t$ from equation \ref{eq_feature_matrix} are matrices which after being normalized continue to the clustering phase of Selective Sharing where the task groups are defined.

\subsection{Factor clustering}

\begin{figure}
	\includegraphics[width=\linewidth]{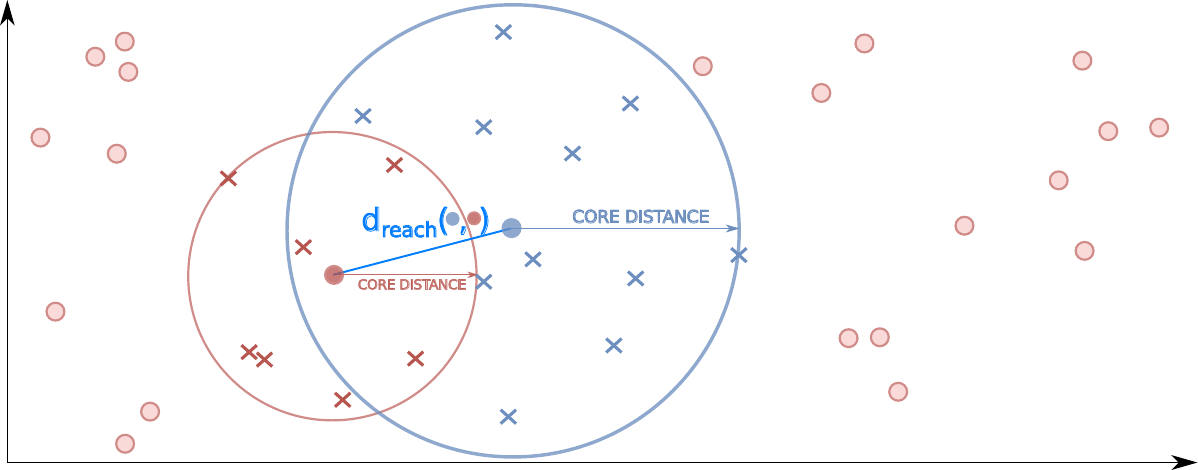}
	\caption{Illustrating the definition of core and mutual reachability distances with two clusters. Core distance we define as the radius of each circle enclosing the factors (marked with X). The mutual reachability distance is defined as the distance between the two centroids according to equation \ref{eq_mutual_reachability_distance}. \label{fig_core_dist}}
\end{figure}

Due to their high dimensionality, density and the critical time period in which clustering occurs, a robust fast high dimensional clustering approach is required. Robust to noise and efficient in handling dense representations, HDBSCAN \cite{mcinnes2017accelerated, campello2013density} is the clustering approach we utilize for our method. It uses a simple distance metric which allows for fast cluster formation and easy selection:

\begin{equation}
		d_{M-k}(C_{t_i},C_{t_j}) = max\left \{  core_k(C_{t_i}), core_k(C_{t_j}), d(C_{t_i},C_{t_j})\right \},
		\label{eq_mutual_reachability_distance}
\end{equation}
\\
where $core_k$ is the distance to the $k-th$ nearest neighboring factor and provides a sort of density estimation, $d(C_{t_i},C_{t_j})$ is the original metric distance between $C_{t_i}$ and $C_{t_j}$ (see Figure \ref{fig_core_dist}). This distance metric is called mutual reachability distance. Under this metric, dense points (with low core distance) remain the same distance from each other whereas sparser points are pushed away to be at least their core distance away from any other point. Its simple definition (eq. \ref{eq_mutual_reachability_distance}) allows for distance metric manipulations so we can effectively alter the task group formation conditions when computing the $R_t$ sets of related tasks.

Clustering is performed over the stacked and normalized gradient factors per task at the end of each epoch. This process repeats until all of the tasks have formed a group with at least one other task, or there have been no changes to the model's architecture for three \footnote{after experimenting with two, three, four and five epochs we empirically determined the number three as most suitable. As clustering is expensive performing it in states where no group formation is possible is redundant.} consecutive epochs. If no stop criteria exists, as training progresses, errors get smaller and gradients sparser, it is expected behavior that all tasks form one group, reverting back to a hard shared MTL approach defying the purpose of grouping in the first place. 

\subsection{Branch merging and recalculation}

When tasks form a group in the clustering phase, it implies that they have similar reactions to the same input, therefore the weights of the layers in their branches should be in a similar state at that point in time. The starting condition of all branches having identical initial states supports this hypothesis. With identical architectures per branch, merging can be performed by regular arithmetic operations in the order of keeping the pairwise maximum, minimum and mean weights, or just keeping the branch whose aggregated loss was the lowest in the newly formed group. For the experimental design we keep the branch with the lowest aggregated loss with its weights and parameters intact.

\section{Experimental Design}

We evaluate Selective Sharing on several problems, both homogeneous and heterogeneous ones. By performing various classification, regression and ranking tasks on five datasets we aim to:
  
\begin{itemize}
	\item Illustrate the reasoning behind the method and the group formation logic both visually and intuitively on simple datasets (MNIST and CIFAR10)
	\item Evaluate group formation handling when optimizing for a very large number of tasks (50) (OmniGlot)
	\item Evaluate performance in ranking problems and compare against classic and current MTL approaches (UCSD-Birds, OmniGlot)
	\item Demonstrate performance in a real large-scale multi-modal MTL problem (OmniArt)
\end{itemize}

\subsection{Datasets and Tasks}

The \textbf{MNIST} \cite{lecun1998gradient} handwritten digit classification problem and \textbf{CIFAR10} \cite{krizhevsky2009learning} are well established benchmarks containing 70K and 60K images. For both experiments we split the data into 50K (MNIST) and 40K (CIFAR10) train, 10K validation and 10K test images spanning 10 target classes. In the CIFAR10 experiment we normalized the images with the training dataset mean. For MNIST we did not use any augmentation or preprocessing steps. As setup in a homogeneous setting we treat the ten target classes as ten binary classification tasks with the standard train/test splits.

\textbf{OmniGlot} \cite{Lake1332} contains 1,623 handwritten from 50 alphabets. Each character is drawn in square regions which we resized to 28x28 pixels and the model is tasked with classifying the alphabet of origin for the input character resulting in a 50 task MTL problem.

\textbf{Caltech-UCSD Birds} dataset \cite{WahCUB_200_2011} provides 11.788 bird images over 200 bird species with 312 binary attribute annotations. For state of the art comparison, we compare on ten target attributes obtained with spectral clustering using the FSIC as the similarity measure \cite{mtl_msd}. For the ten selected attributes, 10 MTL ranking tasks are defined. For each target attribute (ranking task) we rank pairs of images based on the estimated presence of that attribute on 10\% of the dataset. The remaining 80\% is used for training and 10\% used for validation.

\textbf{OmniArt} \cite{Strezoski:2018:OLA:3282485.3273022} provides the data for a comprehensive real large-scale multi-modal MTL problem. This artistic dataset features over 2M data samples and presents a natural heterogeneous MTL problem described by multiple interconnected contextual attributes, making it ideal for testing our method. For the purpose of this experiment we select a subset of OmniArt containing only artworks of the type \textit{painting}. Having a consistent artwork type allows us to better illustrate the connection between the tasks. Our selected subset consists of 133K paintings for which we are using a 80-10-10 split for training, validation and testing and compare against an MTL baseline introduced with the dataset itself \cite{DBLP:journals/corr/abs-1708-00684}.

\begin{table}[]
	\centering
	\caption{Mean performance comparison for MNIST, CIFAR10 and OmniGlot for Selective Sharing, an STL and MTL baseline. Prior knowledge for relatedness between the 50 OmniGlot alphabets was missing so a fair comparison is not possible.}									
	\label{tab_mnist_cifar_omniglot}
	\begin{adjustbox}{max width=\linewidth}
		\begin{tabular}{lccc}
			\hline
			Dataset              					 	& MNIST  			& CIFAR10 		  & OmniGlot     \\
			Metric               					 	& Accuracy (\%) 	& Accuracy (\%)   & Error Rate   \\
			\hline
			STL Baseline \cite{ciresan2011convolutional}& 98.0 			&  92.6		      & 0.34         \\
			MTL Baseline         						& 96.8 			&  90.3 		  & 0.29 \cite{yang2016deep}        \\
			MTL Stitching \cite{misra2016cross}		 	& 98.4 			&  92.1 		  & n/a          \\
			MTL MSD 	  \cite{mtl_msd}	 		 	& 98.7 			&  91.9 		  & n/a          \\
			\rowcolor{Gray}
			\textbf{Selective Sharing - Similarity}  	& \textbf{99.0} 	&  \textbf{92.7} &\textbf{0.24} \\ \hline  
		\end{tabular}
	\end{adjustbox}
\end{table}

\begin{table}[]
	\centering
	\caption{Ranking performance in UCSD Birds. Kendall's Tau correlation scores $x$ 100 $\pm$ std. $x$ 100 are presented. Attributes for which Selective Sharing has superior performance are highlighted in gray.}
	\label{tab_birds}
	\begin{adjustbox}{max width=\linewidth}
		\begin{tabular}{cccccc}
			\hline
			Att. 	  & MTL & RegMTL \cite{evgeniou2004regularized}    & AMTL \cite{pmlr-v48-leeb16}   & MTL-MSD \cite{mtl_msd}    & Ours \\
			\hline
			1         &  51.1$\pm$1.22    & \textbf{60.4}$\pm$3.24 & 59.4$\pm$4.26 & 58.7$\pm$3.69 &         56.1$\pm$1.44         \\
			\rowcolor{Gray}
			2         &  44.3$\pm$0.41    & 54.2$\pm$4.33 & 53.4$\pm$3.09 & 53.9$\pm$2.46 &         \textbf{54.9}$\pm$2.28         \\
			3         &  52.7$\pm$0.27    & \textbf{61.7}$\pm$2.69 & 60.1$\pm$4.44 & 60.1$\pm$1.53 &         61.2$\pm$1.31         \\
			\rowcolor{Gray}
			4         &  49.9$\pm$1.11    & 54.2$\pm$4.17 & 56.5$\pm$1.76 & 54.4$\pm$1.29 &         \textbf{56.1}$\pm$1.25         \\
			\rowcolor{Gray}
			5         &  50.0$\pm$0.70    & 55.0$\pm$3.55 & 51.9$\pm$6.61 & 53.2$\pm$2.09 &         \textbf{55.6}$\pm$1.13         \\
			\rowcolor{Gray}
			6         &  50.0$\pm$0.14    & 54.0$\pm$3.51 & 55.8$\pm$1.09 & 52.3$\pm$1.95 &         \textbf{56.7}$\pm$1.74         \\
			7         &  41.1$\pm$1.31    & 49.0$\pm$10.49& 50.8$\pm$5.46 & \textbf{51.5}$\pm$1.03 &         49.4$\pm$0.99         \\
			\rowcolor{Gray}
			8         &  59.0$\pm$2.63    & 63.0$\pm$2.32 & 63.0$\pm$1.02 & 60.7$\pm$1.91 &         \textbf{63.4}$\pm$1.86         \\
			\rowcolor{Gray}
			9         &  51.4$\pm$0.98    & 53.9$\pm$5.16 & 55.6$\pm$2.66 & 55.7$\pm$1.15 &         \textbf{56.1}$\pm$1.24         \\
			\rowcolor{Gray}
			10        &  51.2$\pm$0.84    & 60.3$\pm$4.21 & 59.5$\pm$6.81 & 61.0$\pm$1.70 &         \textbf{62.2}$\pm$1.83         \\
		\end{tabular}
	\end{adjustbox}
\end{table}

\begin{table*}[t]
	\centering
	\caption{Performance comparison for the OmniArt dataset between a baseline STL approach, a baseline MTL approach and Selective Sharing with Similarity. For every attempt we list the final number of trainable parameters, which for the STL approach is the sum of all the trainable parameters for all tasks. Since the feature extraction is fixed, the parameter count contains only the task specific branches and the hard shared layer.}
	\label{tab_omniart_perf}
	\begin{tabularx}{\textwidth}{lcccccccccc}
		\hline
		Task & AA    & GC    & CCC   & SC   & MC   & STC   & CYE  & \multirow{2}{*}{\# Params} & \multirow{2}{*}{Lock Epoch} & \multirow{2}{*}{\# Branches} \\ 
		Metric              		 	 	  & \multicolumn{6}{c}{\textit{Accuracy (\%)}}  & \textit{MAE (years)} 	       &  &     &  \\ \hline
		STL Baseline        	     		  & 28.0 & 25.0 & \textbf{32.1} & 24.7 & 41.3 & 19.2 & \multicolumn{1}{c}{144.32} & 3,487,664 & N/A & N/A\\
		MTL Baseline                 		  & 31.0 & 26.2 & 27.7 & 23.0 & 42.0 & \textbf{22.1} & \multicolumn{1}{c}{135.43} & 1,630,664  & N/A & 7  \\
		\rowcolor{Gray}
		\textbf{Selective Sharing - Similarity}       & \textbf{33.8} & \textbf{28.0} & 29.0 & \textbf{27.6} & \textbf{44.1} & 21.2 & \multicolumn{1}{c}{\textbf{128.11}} & \textbf{908,264} & 10 & \textbf{3} \\
	\end{tabularx}
\end{table*}

\subsection{Model setup and Training}

For MNIST we define a two layer CNN with 20 5x5 filters in the first layer, and 40 filters of size 5x5 in the second. The convolutions are batch normalized and followed by max pooling layers with stride two. As a feature cutoff point, we apply a 640 unit fully connected (FC) layer. For STL this layer is followed by a sequence of 100 unit and 50 unit FC layers ending in a ten unit softmax output layer \cite{ciresan2011convolutional}. For MTL, the 640 unit hard shared representation is followed by the same classification block from the STL setup, duplicated ten times. We replace the ten-way softmax output with one sigmoid unit per block that performs binary classification. For binary classification we use binary cross-entropy and in STL approaches - categorical cross-entropy as the loss function. 

For CIFAR10 we train a VGG-16 network from scratch with an average pooling cutoff at the end. For STL, the average pooling is followed by 1024 and two 100 dimensional FC layers with a 10-way softmax at the end. For the MTL experiment we rely on the 1024 dimensional FC layer as the hard shared layer and define 10 task specific branches with two 100 unit FC layers and a sigmoid unit for binary classification. 

For Caltech-UCSD Birds 200 (CUB-200), the tasks are posed as ten ranking problems, each ranking pairs of images according to the predicted presence of the attribute in the input images. A pretrained VGG-16 network constitutes the feature extraction unit, followed by a 2048 dimensional FC layer as the hard shared representation. Our 10 ranking blocks have two FC layers of 512 and 256 units followed by an output unit. We train our ranking blocks with a paired ranking loss function:

\begin{equation}
l_{rank}(x_{i},x_{j};f) = max(0,1-(f(x_{i})-f(x_{j})))^2,
\end{equation}

For OmniGlot we train the same three layer CNN from \cite{yang2016deep} with a difference in the classification setup and input dimensions (resized to 28x28px), however, we define 50 tasks and do not make a character level distinction. For the STL experiments we define a FC layer of 300 units, two FC layers of 100 units with 0.2 dropout and a 50 class softmax layer at the end. For the MTL experiments we rest on the same 300 unit hard shared FC layer and define 50 task specific estimators. The task estimators have the STL estimator design, but perform binary classification at the end.

For MNIST, CIFAR, OmniGlot and UCSD-Birds we optimize with Stochastic Gradient Descent with momentum. The MNIST, CIFAR and OmniGlot experiments start with a batch size of 64, learning rate of 0.02 and momentum of 0.5 \cite{ciresan2011convolutional}. The UCSD-Birds experiment optimizes for a batch size of 32, learning rate of 0.01 \cite{evgeniou2004regularized}. All experiments ran for 30 epochs and testing is done with the best performing validation model.

In the OmniArt setup, it is important that we maintain consistent features over all experiments as it is a complex dataset. For this we trained a deep variational autoencoder \cite{DBLP:journals/corr/HouSSQ16a, kingma2013auto} on the training set, which we continue to use as a feature extractor for all STL and MTL experiments. As the estimator, for both the STL and MTL experiments we use the same structure after the hard shared representation layer of 2048 units. The estimators consist of two 300 unit FC layers, 0.2 dropout and softmax for classification or sigmoid for regression outputs. For the STL experiments we train this structure independently for all seven tasks. In the MTL experiments, it comprises the task specific estimators (branches). We define seven tasks of interest, namely artist attribution (AA), genre classification (GC), creation century classification (CCC), school classification (SC), medium classification (MC), style classification (STC) and creation year estimation (CYE). Different from the MNIST, CIFAR10 and OmniGlot datasets, our tasks are not binary classification tasks. Each one has a different number and type of targets. 

\begin{figure}
	\includegraphics[width=\linewidth]{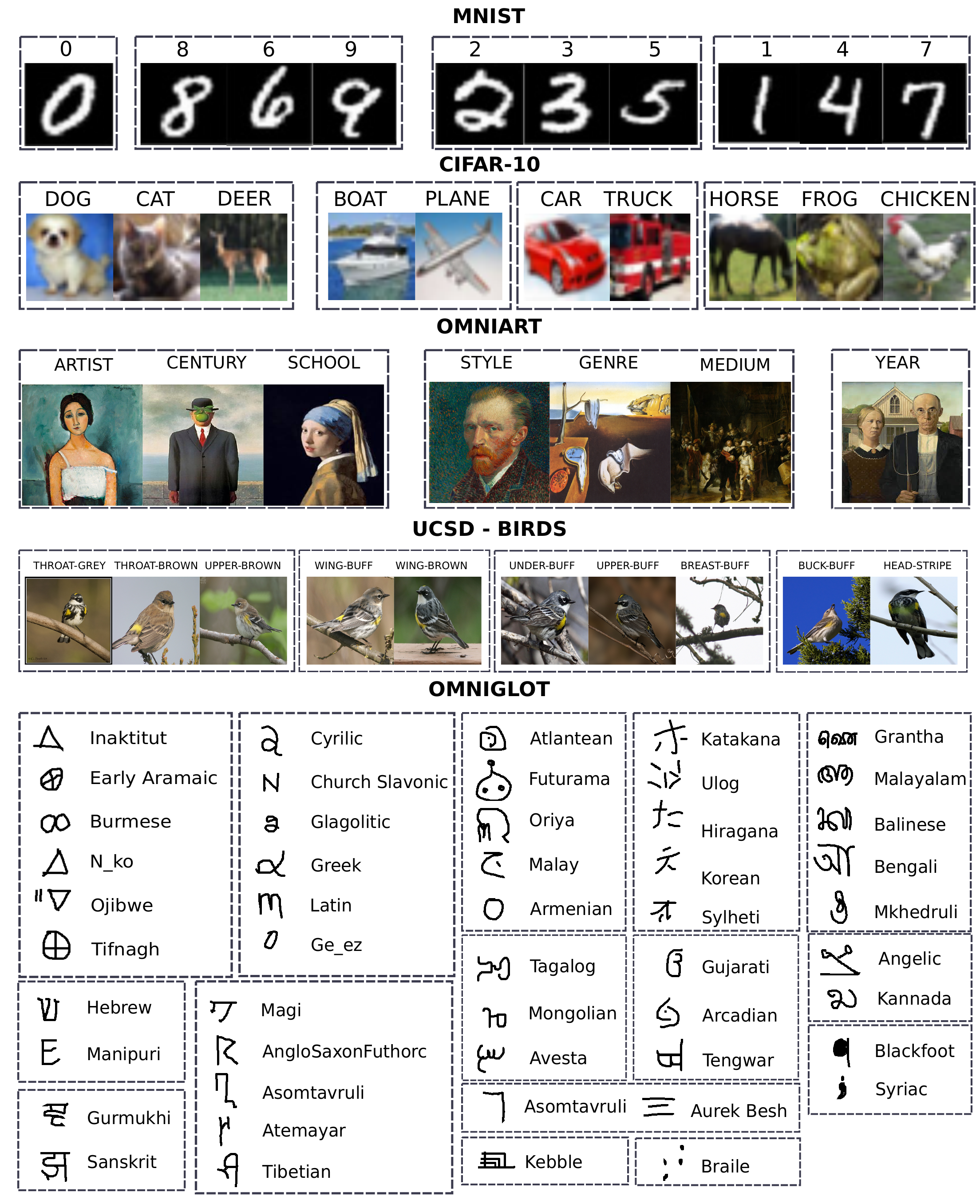}
	\caption{Resulting group formations from applying Selective Sharing on MNIST, CIFAR-10, UCSD Birds, OmniGlot and OmniArt. \label{fig_sharing_options_all}}
\end{figure}

\begin{figure}
	\includegraphics[width=\linewidth]{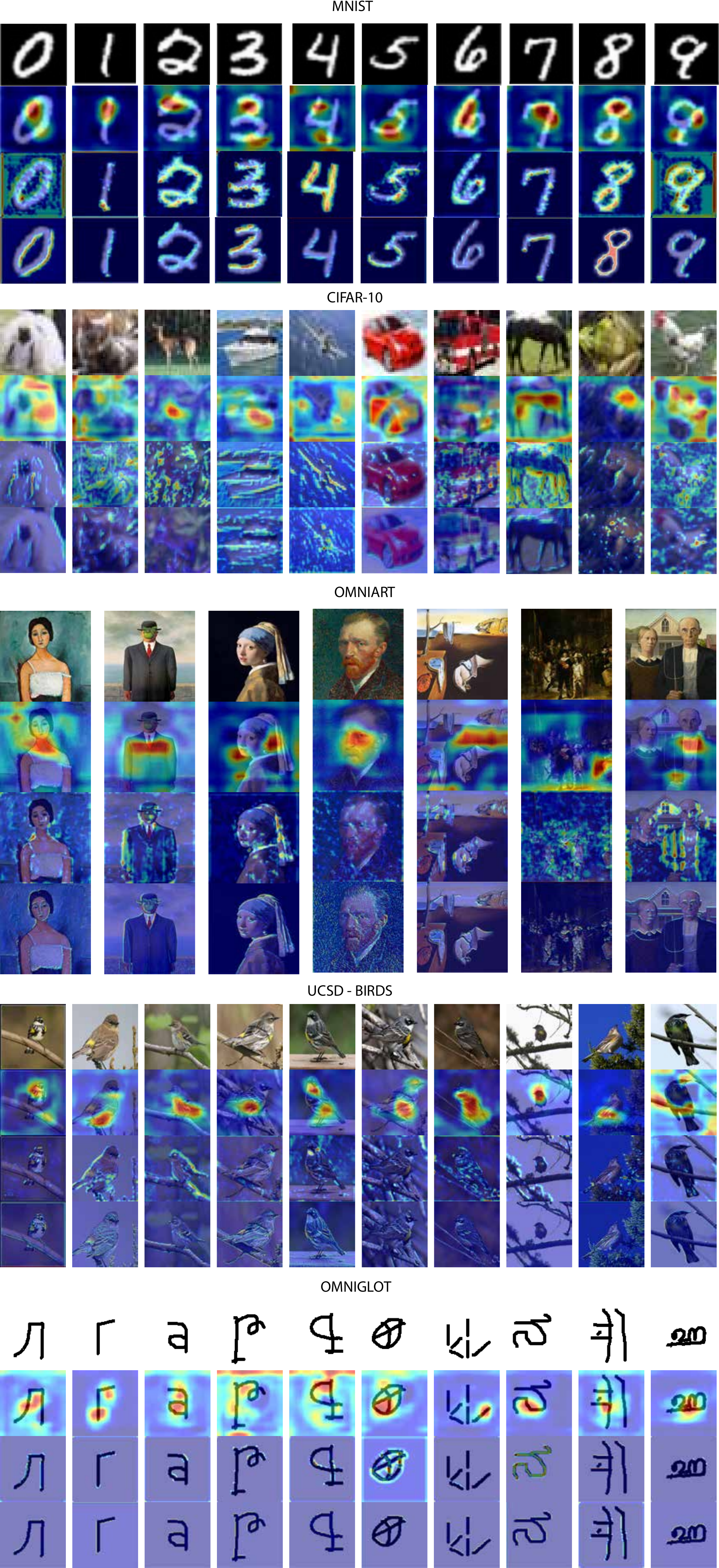}
	\caption{Class activation mappings (CAM) for the correctly predicted class. In Row 1 the input image is displayed, Row 2 shows the CAM in Selective Sharing with Similarity with the grouping presented in Figure \ref{fig_sharing_options_all}. We can observe that features obtained with Selective Sharing are far more general and feathered when it comes to the activation region, making them more general. Row 3 shows the CAM for the STL baseline and Row 4 the CAM for the MTL baseline. \label{fig_cam_figs}}
\end{figure}

For the OmniArt experiments we use the Adam optimizer \cite{kingma2014adam} with learning rate of 0.001, batch size of 100 and train for 50 epochs. Unless specified, all nonterminal FC units use the ReLU \cite{nair2010rectified} activation function. Testing is done with the best performing validation models. 

\subsection{Results and Discussion}

Our experimental design explores multiple sharing paradigms for the five datasets. The reported results show the performance with Selective Sharing with Similarity as the best performing approach with Figure \ref{fig_sharing_options_all} showing the formed groups for which performance is shown in Tables \ref{tab_mnist_cifar_omniglot} and \ref{tab_omniart_perf} \footnote{Group formations, results with all sharing criteria and training details are available in the supplementary material.}.

Sharing with similarity yields an accuracy of 99.0\% for MNIST. For CIFAR10, sharing with similarity produced an average per class (per task) accuracy of 92.7\% improving upon both the STL and MTL baseline approaches. Selective Sharing with similarity (expectedly) yields the best results (MNIST, CIFAR and OmniGlot), partly because similar letter and number types share the same constructional primitives (see Figure \ref{fig_cam_figs}). In that way, learning similar alphabets together is expected to improve the generalization capabilities of the model. Using a superior feature extraction method, data augmentation and parameter tuning can improve the performance in both tasks, but is beyond the purpose of the experiments.

For the UCSD-Birds homogeneous ranking problem we compare against a vanilla MTL model, two strong baselines \cite{evgeniou2004regularized, pmlr-v48-leeb16} and one state-of-the-art approach \cite{mtl_msd} in ranking the images containing ten selected bird attributes. Selective Sharing with similarity outperforms all other approaches in seven out of ten tasks. 

Sharing with similarity produced the best results for the OmniArt dataset overall, while sharing with between cluster variance mostly boosted performance for AA and STC tasks. The consistent improvement of sharing with similarity over the MTL baseline can be observed in Table \ref{tab_omniart_perf}. E.g. the Dutch school of painting was active in the 1400s - late 1500s. Most of the artists from this period in our OmniArt subset are Dutch masters, which implies that knowing the period and school narrows down the list of possible artists significantly. We could have easily discovered this relation by mining the data itself, but the quadruple correlation in sharing with variance between artists, creation periods, schools and styles is quite intricate and model driven. 

A qualitative observation from the activation region analysis is that the representations learned with selective sharing are more general and feathered than their STL or MTL counterparts. In addition, as visible the second rows of Figure \ref{fig_cam_figs}, other than the larger activation span we can observe a similarity between the class activation regions for tasks that formed groups. For example the group formed between '2','3' and '5' in the MNIST tasks exhibits a consistent activation on the curve of the digit in all three tasks. A similar activation can be observed in the below neckline and shoulder area in the Artist, Century, School group in the OmniArt analysis.

\section{Conclusions}

Selective Sharing is a multi-task learning approach for images in context that allows exploiting task relatedness for group formation without any predefined intertask dependency. Group formation is performed using conditional statements on distances between clusters of task specific gradient factors. This grouping procedure implicitly reduces the trainable parameter space dimensionality and boosts predictive performance for related tasks (or contextual attributes) as a result (see Tables \ref{tab_mnist_cifar_omniglot}, \ref{tab_birds}, \ref{tab_omniart_perf}). Moreover, Selective Sharing is modality invariant, making it applicable in every multimedia scenario where a supervision signal and multiple targets are available. As such, applied on a wide range of MTL problems, Selective Sharing adjusts the representation learning process to implicitly exploit the entanglement between the contextual attributes and supports learning robust shared representations.

On a general note, experimental results on images in context show that forming groups of tasks over which we build a mutual representation is beneficial to the overall learning process. Figure \ref{fig_cam_figs} illustrates that the features we obtain are more robust and general in comparison to STL and MTL baselines. Furthermore, the way a model defines its groups, affects the performance in specific ways. E.g. sharing with similarity can improve tightly  interconnected tasks, but sharing with variance can improve the overall performance as the latent data representation is built with respect to more contextual attributes that correlate and hold exclusive data insight. Compared to conventional matrix driven MTL approaches, Selective Sharing provides an additional degree of freedom in modeling the tasks at hand, as between task relations need not be studied prior to the classification architecture design. Moreover, the intuition and logic behind the approach is simple and easy to grasp without the need for complex mathematical or statistical apparatus, making Selective Sharing a versatile and useful approach to MTL. Conclusively, sharing is a virtue and knowing who to share with - is awareness. We attempt to teach our models how to do both.